\documentclass[lettersize,journal]{IEEEtran}
\usepackage{amsmath,amsfonts}
\usepackage{algorithmic}
\usepackage{algorithm}
\usepackage{array}
\usepackage[caption=false,font=normalsize,labelfont=sf,textfont=sf]{subfig}
\usepackage{textcomp}
\usepackage{stfloats}
\usepackage{url}
\usepackage{hyperref}
\hypersetup{hidelinks}
\usepackage{verbatim}
\usepackage{graphicx}
\usepackage{booktabs}
\usepackage{cite}
\usepackage{inconsolata}
\usepackage{bm}
\usepackage{multirow}
\begin{document}

\title{CDFKD-MFS: Collaborative Data-free Knowledge Distillation via Multi-level Feature Sharing}

\author{
    Zhiwei~Hao, Yong~Luo,~\IEEEmembership{Member,~IEEE,} Zhi~Wang~\IEEEmembership{Member,~IEEE,} Han~Hu,~\IEEEmembership{Member,~IEEE,} and~Jianping~An,~\IEEEmembership{Member,~IEEE,}
    \thanks{Corresponding author: Han Hu.}
    \thanks{Zhiwei Hao, Han Hu, and Jianping An are with the School of Information and Electronics, Beijing Institute of Technology, Beijing 100081, China (e-mail: haozhw@bit.edu.cn, hhu@bit.edu.cn, an@bit.edu.cn).}
    \thanks{Yong Luo is with the School of Computer Science, Wuhan University, Wuhan 430072, China (e-mail: luoyong@whu.edu.cn).}
    \thanks{Zhi Wang is with Shenzhen International Graduate School, Tsinghua University, Shenzhen 518055, China (e-mail: wangzhi@sz.tsinghua.edu.cn).}
    \thanks{The code is available at \url{https://github.com/Hao840/CDFKD-MFS}}
}

\markboth{A SUBMISSION TO IEEE TRANSACTIONS ON MULTIMEDIA}%
{Hao \MakeLowercase{\textit{et al.}}: A submission to Ieee Transactions on Multimedia}


\maketitle

\begin{abstract}
    Recently, the compression and deployment of powerful deep neural networks (DNNs) on resource-limited edge devices to provide intelligent services have become attractive tasks.
    Although knowledge distillation (KD) is a feasible solution for compression, its requirement on the original dataset raises privacy concerns.
    In addition, it is common to integrate multiple pretrained models to achieve satisfactory performance.
    How to compress multiple models into a tiny model is challenging, especially when the original data are unavailable.
    To tackle this challenge, we propose a framework termed collaborative data-free knowledge distillation via multi-level feature sharing (CDFKD-MFS), which consists of a multi-header student module, an asymmetric adversarial data-free KD module, and an attention-based aggregation module.
    In this framework, the student model equipped with a multi-level feature-sharing structure learns from multiple teacher models and is trained together with a generator in an asymmetric adversarial manner.
    When some real samples are available, the attention module adaptively aggregates predictions of the student headers, which can further improve performance.
    We conduct extensive experiments on three popular computer visual datasets.
    In particular, compared with the most competitive alternative, the accuracy of the proposed framework is 1.18\% higher on the CIFAR-100 dataset, 1.67\% higher on the Caltech-101 dataset, and 2.99\% higher on the mini-ImageNet dataset. 
\end{abstract}



\section{Introduction}
\label{sec:intro}
\IEEEPARstart{D}{uring} the past few years, deep neural networks (DNNs) have become one of the most influential models in machine learning, and with the developments of the 5G wireless network and artificial intelligence of things (AIoT), equipping resource-constrained edge devices with DNN-based intelligent applications has become appealing.
Unfortunately, modern DNNs are heavily overparameterized, requiring tremendous storage and computing resources for inference.
For example, ResNet-50 \cite{resnet}, a typical DNN architecture for computer vision, occupies more than 100 megabytes of ROM for storage and requires more than 4 giga FLOPs for inference.
Such resource greedy DNNs make deployment challenging.
Knowledge distillation (KD) \cite{kd} provides a feasible solution by compressing a large pretrained teacher model (teacher) into a tiny student model (student).
However, the dataset for training the large model, which is indispensable in KD, is sometimes unavailable due to the high transmitting overhead or privacy concerns.
Although some data-free KD approaches have been proposed \cite{dfad,adi}, they are not suitable for multi-model scenarios, where multiple models are integrated in an ensemble to achieve better performance.
How to effectively compress multiple models by KD in a data-free manner is still an open problem. 

There are some obstacles to achieving a multi-teacher data-free KD.
First, the student may be confused from the knowledge provided by different teachers.
Since the benefit of an ensemble model usually comes from its diverse feature representations learned by multiple models \cite{DBLP:journals/corr/abs-2012-09816}, a student with a linear-structure that is directly trained using these models may not be able to sufficiently preserve diversity, which results in performance degradation.
Second, conducting KD requires careful design.
Directly regarding the ensemble as a single model to apply single-teacher KD is convenient, but this method may ignore the diversity of different teachers and lead to a suboptimal result.
Finally, how to effectively leverage available samples from the original dataset for data-free KD for better performance is challenging.

Some works designed multi-branch student architectures to avoid knowledge confusion when the original dataset is available.
For example, Tran \textit{et al.} \cite{DBLP:journals/corr/abs-2001-04694} proposed a tree-like student architecture for multi-teacher KD, and Liu \textit{et al.} \cite{DBLP:conf/cvpr/LiuJD19} designed a multi-task attention network to learn multiple tasks simultaneously.
A few works studied how to achieve KD with multiple teachers when the original data are unavailable.
Ye \textit{et al.} \cite{DBLP:conf/cvpr/YeJWGS20} proposed a group-stack dual-generative adversarial network (GAN) architecture.
However, their work mainly focused on the knowledge amalgamation (KA) problem. 
Their model aimed to amalgamate functions of multiple models in one model and little attention was paid to the size of the target model.
Specifically, a linear network was trained as a backbone and the last several layers of teachers was grafted the onto the backbone.
This approach achieved KA but did not compress the teacher branches.
Moreover, the backbone of the model is a concatenation of multiple separately trained parts, which may not reach the global optima.
For the second challenge, generator-based approaches \cite{DBLP:conf/iccv/ChenW0YLSXX019,dfad} provide a favorable solution for the single-teacher scenario.
Furthermore, in data-free quantization, where the teacher and the student have the same architecture, intermediate feature supervision has been introduced \cite{DBLP:conf/cvpr/LiuZW21a}.
However, it is still desirable to design an approach to conduct data-free KD on multiple teachers and a student with a different architecture.
Finally, using real samples in data-free KD has not been studied.
Because a potential domain gap exists between real and generated samples, directly combining them to train a student is inappropriate.

To tackle these issues, we propose a collaborative data-free knowledge distillation framework with a multi-level feature-sharing (CDFKD-MFS) student architecture.
This framework is an extension of our previous conference work \cite{cdfkd}.
Our new CDFKD-MFS framework consists of three parts: a multi-header student network module, an asymmetric adversarial data-free KD module, and an attention-based prediction aggregation module.
First, the student is composed of a backbone network and multiple lightweight header networks and takes the multi-level feature of the backbone as input.
This architecture alleviates the knowledge confusion problem in multi-teacher KD by decoupling parameters for learning shared knowledge and teacher-specific knowledge.
Then, the student and a data generator are adversarially trained based on the distillation loss.
In addation, the generator is guided by statistics in the batch normalization layers of teachers, and the student is trained using an ensemble distillation loss and a feature distillation loss.
Finally, for cases when some real samples are available, an attention module is designed to aggregate predictions of multiple headers in the student.
We evaluate the proposed framework on three datasets with different sample sizes: CIFAR-100 \cite{cifar}, Caltech-101 \cite{caltech101}, and mini-ImageNet \cite{miniimagenet}.
The results show that our method achieves at least $1.18\%$, $1.67\%$, and $2.99\%$ performance improvements compared to the next best model on these datasets, respectively.
We summarize our contributions as follows:
\begin{itemize}
    \item We propose a data-free KD framework for multi-model compression, where multiple pretrained teachers collaborate to obtain a better student.
    \item We design a multi-header student equipped with a multi-level feature-sharing architecture, where its backbone learns shared knowledge, and each header learns from a specific teacher.
    \item We design an asymmetric adversarial data-free KD algorithm, where a data generator and the student are trained in an asymmetric adversarial way.
    \item We develop an attention-based prediction aggregation module, which can further improve performance when a few samples from the original dataset are available.
\end{itemize}

The rest of this paper is organized as follows.
Section \ref{sec:ralated_works} introduces existing works related to KD, branchy networks, and attention.
Section \ref{sec:method} provides details of the proposed framework.
We evaluate the framework in Section \ref{sec:exp} and finally summarize our work in Section \ref{sec:conclusion}.

\section{Related Works}
\label{sec:ralated_works}

\subsection{Data-required KD}
\label{subsec:rw_kd}
Knowledge distillation, which is a practical approach for compressing a large pretrained teacher model into a tiny student model, was first proposed by Hinton \textit{et al.} \cite{kd}.
The original KD method takes predictions of the teacher model as soft labels to train the tiny student model from a sketch.
Based on this framework, some works simulates soft labels by regularizing \cite{DBLP:journals/corr/abs-1908-05474} or label smoothing \cite{DBLP:conf/nips/MullerKH19,DBLP:conf/iclr/KimK17}.
In addation to the output layer, there are also some approaches for transferring the knowledge in intermediate layers of the teacher.
Based on this consideration, FitNet \cite{DBLP:journals/corr/RomeroBKCGB14} and some follow-up works \cite{DBLP:conf/eccv/WangFLWLM20,DBLP:conf/eccv/XuRLG20} adopted feature matching losses to utilize intermediate knowledge.
Moreover, intermediate-layer supervision can also be achieved through attention map matching \cite{DBLP:conf/iclr/ZagoruykoK17}, feature distribution matching \cite{DBLP:conf/eccv/PassalisT18}, and cross-layer knowledge transferring \cite{DBLP:conf/aaai/ChenMZWWF021}.

Multiple pretrained teacher models can provide diverse knowledge and improve the performance of the student \cite{DBLP:journals/corr/abs-2012-09816}.
Hinton \textit{et al.} \cite{kd} proposed using the average of teacher prediction as the supervising signal. 
You \textit{et al.} \cite{DBLP:conf/kdd/YouX0T17} and Park \textit{et al.} \cite{DBLP:conf/ecai/ParkK20} introduced feature supervision in multi-teacher KD.
To better transfer knowledge from multiple teachers, some works added additional branches to the student \cite{DBLP:conf/ecai/AsifTH20, DBLP:journals/corr/abs-2001-04694}.
However, these branches were directly grafted onto the student without careful consideration, which may not be sufficient for preserving the diverse knowledge of teachers.
Hence, multi-teacher KD requires a student to learn diverse knowledge from multiple teachers while remaining lightweight.
To tackle this problem, we design a multi-header student with a multi-level feature-sharing structure to learn from multiple teachers.

\subsection{Data-free KD}
\label{subsec:rw_dfkd}
Data-free KD aims to achieve knowledge transfer in the absence of the original dataset.
A direct idea is to use other datasets as a substitute \cite{DBLP:conf/wacv/NayakMC21,DBLP:conf/cvpr/Chen00LX0021}, or synthesize a dataset to implement KD.
Based on whether a generator is adopted, we categorize approaches using synthesized data into two classes: optimization-based and generator-based approaches.

Optimization-based approaches directly optimize a random initialized sample to make it suitable for KD.
Lopes \textit{et al.} \cite{DBLP:journals/corr/abs-1710-07535} reconstructed the original dataset using metadata collected during teacher model training.
Nayak \textit{et al.} \cite{DBLP:conf/icml/NayakMSRC19} produced samples by modeling the softmax space of the original dataset. 
Recently, statistics in batch normalization (BN) layers of the pretrained teacher model have been adopted to assist data generation \cite{adi, DBLP:conf/cvpr/HaroushHHS20}.
More realistic substitute samples can be obtained with this information.
Fang \textit{et al.} \cite{cmi} further introduced contrastive learning to generate more diverse samples.
However, optimization-based approaches optimize each sample separately, which means that the resource consumption is linearly related to the amount of required data.
Moreover, although the scenario is assumed to be data-free, most of these approaches require the mean and variance (not those in the BN layers) of the original dataset for data generation.

Generator-based approaches adopt a generator to synthesize the substitute dataset, which is usually trained to output samples with given labels \cite{DBLP:conf/nips/YooCKK19,DBLP:journals/corr/abs-2012-05578}, high teacher activation, or low prediction entropy \cite{DBLP:conf/iccv/ChenW0YLSXX019}.
Adversarial distillation \cite{dfad,DBLP:conf/nips/MicaelliS19} is an efficient scheme that trains the generator and the student together by a min-max game.
Fang \textit{et al.} \cite{DBLP:conf/eccv/XuLZLCLT20} and Choi \textit{et al.} \cite{dfq} combined adversarial distillation and a BN statistics constraint for data-free quantization.
Liu \textit{et al.} \cite{DBLP:conf/cvpr/LiuZW21a} designed a channel relation map matching loss to further improve performance.
Only a few works have studied the multi-teacher data-free KD problem.
Ye \textit{et al.} \cite{DBLP:conf/cvpr/YeJWGS20} designed a group-stack dual-GAN student architecture to amalgamate different knowledge.
Mainly aiming at amalgamation, their student contains layers from the teacher networks, causing inefficient compression.
Moreover, their student architecture is highly related to the generator, which may limit its flexibility, and the student consisting of separate trained blocks may be suboptimal.
Hao \textit{et al.} \cite{dfed} used the adversarial distillation framework to achieve a multi-teacher data-free KD, but the use of a single-header student may confuse diverse knowledge in multiple teacher models.
To relieve this confusion and achieve flexible knowledge transfer, we design a multi-header student and an asymmetric adversarial data-free KD algorithm, which is end-to-end.

Our previous conference work \cite{cdfkd} performed data-free KD with multiple pretrained teacher models under a proposed framework named CDFKD.
The CDFKD framework consists of multiple teacher models, a student model with multiple output headers, and a generator model.
In this framework, the generator produces substitute samples, while the student learns from multiple teachers with these samples, and each teacher is assigned to a corresponding student header.
Moreover, when some samples from the original dataset are available, a simple attention module is trained to aggregate predictions given by student headers.

\begin{figure*}
    \centering
    \includegraphics[width=0.9\textwidth]{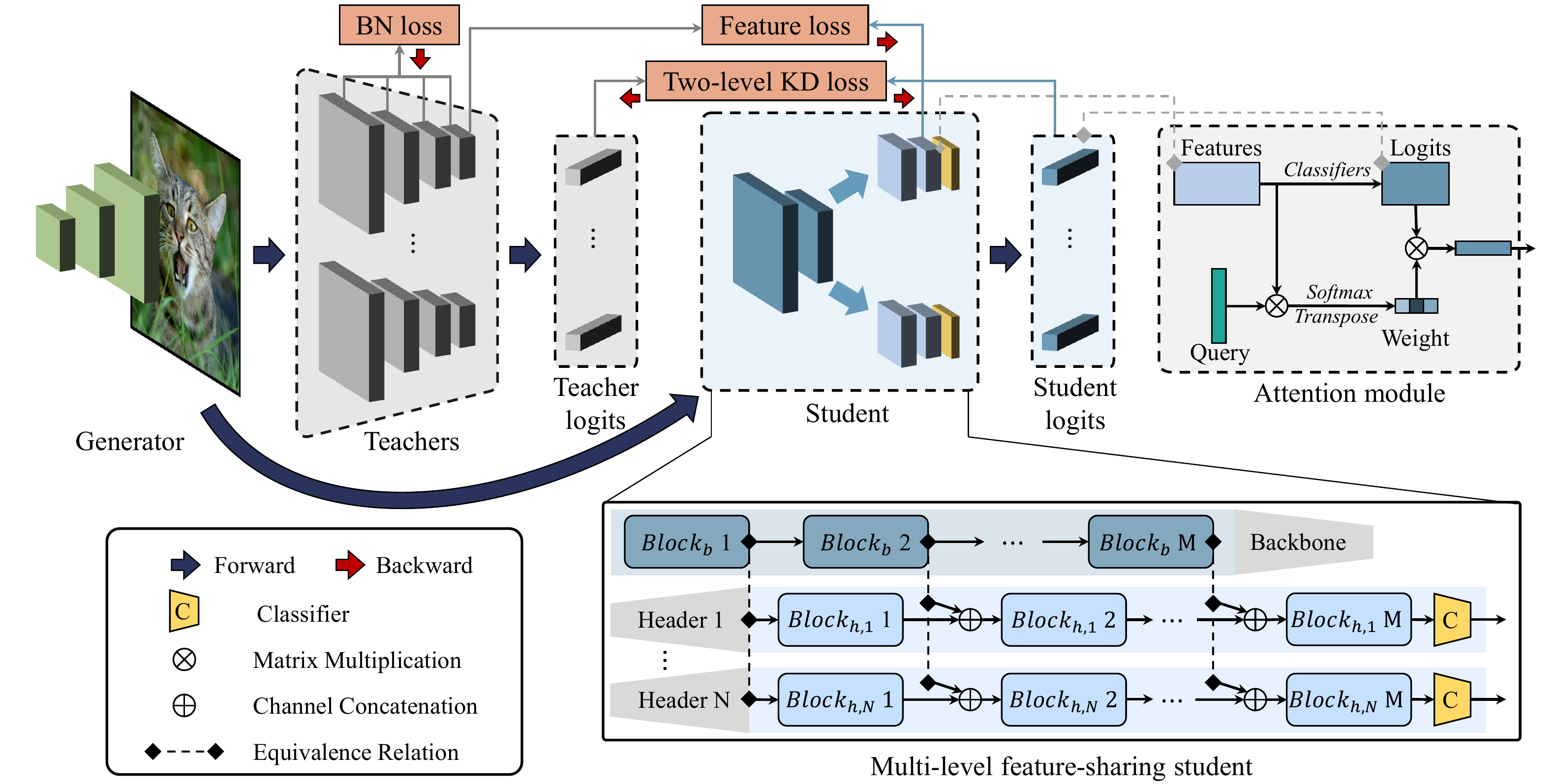}
    \caption{Overview of the CDFKD-MFS framework. This framework consists of multiple pretrained teachers, a data generator, a multi-header student, and an attention module. In this framework, the generator produces samples to substitute the original data, and the student incorporated with multi-level feature sharing learns from teachers using these samples. When some real samples in the original dataset are available, the attention module adaptively aggregates predictions of student headers.}
    \label{fig:overview}
\end{figure*}

\subsection{Decoupling and Aggregation of Knowledge}
\label{subsec:rw_others}

To learn from multiple teachers and preserve the diversity, the student needs to decouple the knowledge in the teacher ensemble and learn with decoupled parameters with a branchy network architecture.
Such designs were common in early-exiting \cite{DBLP:journals/corr/abs-1709-01686} inference and multi-task learning.
The multi-task learning problem, where early works adopted a tree-like student architecture as a solution \cite{DBLP:conf/aaai/HanZSX14}, bears a closer resemblance to our work.
Some later approaches decoupled the network into task-shared and task-specific parts.
Misra \textit{et al.} \cite{DBLP:conf/cvpr/MisraSGH16} proposed a cross-stitch unit to combine activations of the two parts.
Liu \textit{et al.} \cite{DBLP:conf/cvpr/LiuJD19} proposed a multi-task attention network to achieve the decoupling of knowledge.
The student designed for multi-teacher KD in our study was inspired by this work.

Aggregating predictions of the independent branches benefits from the ensemble method.
We achieve aggregation with the attention mechanism, which enhances some parts of the input information while ignoring extraneous parts.
The attention mechanism has been combined with DNNs to learn better feature representation \cite{DBLP:conf/eccv/WooPLK18,DBLP:conf/cvpr/HuSS18}.
In addation to these simple attention mechanisms, more complex forms such as self-attention networks \cite{DBLP:conf/nips/VaswaniSPUJGKP17} and graph attention networks \cite{DBLP:conf/iclr/VelickovicCCRLB18} have also achieved outstanding performance.
In our work, we adopt key-value attention to aggregate predictions of student headers when a few real samples are given.

\section{Method}
\label{sec:method}
To facilitate data-free KD to compress multiple pretrained models, we propose a framework termed CDFKD-MFS. 
Our framework can integrate the knowledge of multiple models into a tiny model while preserving diverse multi-model knowledge.

\textbf{An overview} of the proposed CDFKD-MFS framework is shown in Fig. \ref{fig:overview}.
This framework contains multiple teacher models, a generator, a multi-header student model, and an attentional aggregation module.
Teacher models are pretrained on the original dataset.
During data-free KD, the generator produces samples to substitute the original data.
Given the generated samples, the student uses its backbone network to learn shared knowledge and its header branches to learn from specific teachers.
In the student model, multi-level feature sharing is incorporated to reduce the branchy model size while providing satisfactory performance.
Moreover, when some real samples from the original dataset are available, the attention module aggregates predictions of student headers to further improve performance.

In the following part of this section, we first introduce some background knowledge of data-free KD. 
Then we depict details of our CDFKD-MFS framework.

\subsection{Background}
\label{subsec:method_background}
Performing data-free KD requires a substitute for the original dataset.
Existing approaches obtain this substitute using either generator-based or optimization-based methods.
Generator-based methods train a generator network to produce substitute samples, while optimization-based methods directly optimize Gaussian noise for desired samples.
Because the proposed framework is generator-based, we introduce some background knowledge and universal techniques of generator-based data-free KD, including KD, adversarial distillation, and BN statistics constraint.

\textbf{KD} is a technique for compressing a large pretrained model into a tiny one \cite{kd}.
During training, KD uses the predictions of the pretrained model, called teacher $T$, to guide the learning of a tiny student model $S$.
Denoting the parameters of the student as $\bm{\theta_s}$, the above procedure is formulated as:
$$
    \min_{\bm{\theta_s}} \mathbf{E}_{\bm{x} \in \mathcal{X}} [\mathcal{D}_{KL}(S_{\bm{\theta_s}}(\bm{x}), T(\bm{x})) + \lambda\mathcal{D}_{CE}(S_{\bm{\theta_s}}(\bm{x}), y)],
$$
where $\mathcal{D}_{KL}$ and $\mathcal{D}_{CE}$ denote the KL-divergence and cross-entropy distance, respectively, $y$ is the ground-truth label of input sample $\bm{x}$, and $\lambda$ is a hyperparameter balancing the two objectives.
In practice, we found that simply setting $\lambda$ to $0$ has little influence on student performance, so we adopt this more straightforward KD form in our study.
However, the vanilla KD approach can only work in data-available scenarios. 
When the original dataset is unavailable, adversarial data-free distillation is an ideal solution.

\textbf{Adversarial distillation} is a typical training scheme for generator-based data-free KD approaches \cite{DBLP:conf/nips/MicaelliS19}.
When the original dataset is unavailable, the core problem is to find a substitute for the dataset.
In adversarial distillation, this substitution is obtained by training a data generator, and the substitute data generation runs together with KD adversarially.
Adversarial training is initially used in generative adversarial networks, where a generator network produces samples and a discriminator network distinguishes generated samples from real samples \cite{gan}.
The two networks are trained successively in each iteration.
However, the discriminator trained with the original dataset cannot be directly adopted in the case of missing data.
Adversarial distillation regards the teacher and the student together as a proxy discriminator and trains it with a negative KL divergence loss.
Mathematically, the adversarial distillation procedure can be represented as:
$$
    \min_{\bm{\theta_s}} \max_{\bm{\theta_g}} \mathbf{E}_{\bm{z} \sim \mathcal{N}(0,\bm{I})}[\mathcal{D}_{KL}(S_{\bm{\theta_s}}(G_{\bm{\theta_g}}(\bm{z})),T(G_{\bm{\theta_g}}(\bm{z})))],
$$
where $\bm{\theta_g}$ represents the parameters of generator $G$ and $\bm{z}$ is a random vector sampled from a multivariate normal distribution.
This procedure can be intuitively interpreted as follows:
in each iteration, the generator produces samples on which the student cannot output predictions similar to the teacher.
Simultaneously, the student learns from the teacher using these samples.
When the training converges, the student can learn most teacher knowledge.
Hence, the quality of the generated samples significantly influences the performance of the student, and how to generate samples containing more meaningful information plays a crucial role.

\textbf{BN statistics constraint} has been proven helpful in sample generation for data-free KD \cite{adi, dfq}.
BN layers can reduce the internal covariate shifts when data forward propagate through neural networks, which makes the training of deeper networks achievable \cite{bn}.
In data-free KD, these layers can also reduce the domain shift of generated samples.
Specifically, during training, BN layers of the teacher model compute the mean and variance of all the data passing through them.
By constraining activations of generated samples to match the precollected mean and variance on these layers, the generator can produce samples following a distribution close to the original dataset.
Indexing BN layers with $i$ and denoting the mean and variance of the $i$-th BN layer as $\mu_i$ and $\sigma_i$, the BN statistic constraint is given by:
$$
    \min_{\bm{\theta_g}}\mathbf{E}_{\bm{z} \sim \mathcal{N}(0,\bm{I}),i}[||\hat{\mu}_i(G_{\bm{\theta_g}}(\bm{z})) - \mu_i||_2 + ||\hat{\sigma}_i(G_{\bm{\theta_g}}(\bm{z})) - \sigma_i||_2],
$$
where $\hat{\mu}_i(G_{\bm{\theta_g}}(\bm{z}))$ and $\hat{\sigma}_i(G_{\bm{\theta_g}}(\bm{z}))$ are the mean and variance of activations at the $i$-th BN layer when the teacher takes $G_{\bm{\theta_g}}(\bm{z})$ as input, respectively.

Although remarkable performance was achieved, almost all previous works assumed that only one pretrained teacher model exists without considering multi-teacher scenarios.
We remedy this drawback and propose the CDFKD-MFS framework, which is suitable for multi-teacher data-free KD.

\subsection{Multi-header Student Architecture}
\label{subsec:method_student}
We first introduce a multi-header network architecture, which is adopted as the student in our framework.
As there are multiple pretrained teachers, how to preserve their diverse knowledge into a single model requires careful consideration.
A possible solution is to attach multiple output headers to a student and let each header learn from an individual teacher.
Our previous work attached output headers to the student in a tree-like manner.
More specifically, all headers are directly concatenated at the end of the backbone of the student.
However, this simple architecture is insufficient to learn the abundant, diverse knowledge of multiple teachers and results in performance degradation, especially on some complex datasets, as shown in Section \ref{subsec:exp_cc}. 
To remedy this drawback, we propose a multi-header student with multi-level feature sharing.

The lower right part of Fig. \ref{fig:overview} illustrates the details of the proposed student architecture, which consists of two components: a shared backbone network and multiple header networks.
There are the same number of headers and pretrained teachers.
We denote the number by $N$.
We split the backbone network into several blocks according to where the intermediate features are shared with the headers and assume the number of feature-sharing points is $M$.
Each header also consists of $M$ blocks, which correspond to blocks in the backbone.
The backbone can be any off-the-shelf network architecture, e.g., ResNet \cite{resnet}.
As the number of headers is related to the number of pretrained teachers, headers should be lightweight enough to avoid consuming too many computing resources when performing inference.
Hence, we adopt the depthwise separable convolution \cite{sepconv} to build lightweight headers.
Table \ref{tab:sepconv} provides the architecture of one student header block, where \texttt{Conv($C_{1},C_{2},k$)} denotes a convolutional layer with a $k \times k$ kernel, which takes $C_{1}$ channels feature map as input and output $C_{2}$ channels feature map.

\begin{table}[t]
    \caption{Network Architecture of One Student Header Block.}
    \begin{center}
        \begin{tabular}{c}
        \toprule
        \texttt{Conv($C_{in},C_{in},3$), Conv($C_{in},C_{in},1$), BatchNorm, ReLU}\\
        \texttt{Conv($C_{in},C_{in},3$), Conv($C_{in},C_{out},1$), BatchNorm, ReLU}\\
        \bottomrule
        \end{tabular}
        \label{tab:sepconv}
    \end{center}
\end{table}

The student takes samples produced by the generator as inputs.
During forward propagation, its backbone network performs the same as an ordinary DNN, and each header network takes multi-level intermediate features of the backbone network as input.
For all headers, the input of their first block is the output of the first backbone block.
Then, we take the $i$-th header and its $j$-th ($j>1$) block as the example to explain how other header blocks work.
When the forward propagation of the $(j-1)$-th backbone block and the $(j-1)$-th block of the $i$-th header is finished, their output feature maps are then concatenated at the channel dimension and fed into the $j$-th block.
Once the inference of the last block in a header is finished, the output feature is sent to the corresponding classifier to obtain the final prediction.
The predictions of all headers can be aggregated by taking the average or other ensemble approaches.
Similar network architecture designs can be found in multi-task learning \cite{DBLP:journals/corr/RusuRDSKKPH16, DBLP:conf/cvpr/LiuJD19}, which can alleviate the interference caused by learning different tasks simultaneously.
In the proposed architecture, the backbone learns the knowledge shared by all headers, and each header learns from a specific teacher.
This knowledge decoupling architecture can preserve the diverse knowledge of multiple teachers in a single student network and achieve better performance in comparison with a single-header student architecture, which will be shown in Section \ref{subsec:exp_cc}.

\subsection{Asymmetric Adversarial Data-free KD}
\label{subsec:loss}
After the designation of the multi-header student, we introduce the proposed multi-teacher data-free KD approach.
We name this approach asymmetric adversarial data-free KD because we adopt the adversarial distillation framework, but the objective functions of the generator and the student are not entirely the same.
Specifically, for the generator, its objective function has a BN statistics constraint term, and for the student, feature supervision from teachers is introduced.
Next, we first introduce the three loss terms illustrated in Fig. \ref{fig:overview}, and then introduce the objection function for training the generator and the student.
Finally, the optimization is introduced.

\textbf{The BN loss} is a constraint acting on the mean and variance of activations at the BN layers of all pretrained teachers.
It requires activations of generated samples to have statistics similar to those of samples in the original dataset.
In the multi-teacher scenario, we assume there are $N$ teachers in total and index them with $n$.
Thus, we can extend the BN statistics constraint in Section \ref{subsec:method_background} to the multi-teacher setting:
\begin{equation}
    \begin{aligned}
        \mathcal{L}_{bn}(\bm{\theta_g})=\frac{1}{N}\sum_{n=1}^{N} \mathbf{E}_{\bm{z} \sim \mathcal{N}(0,\bm{I}),i \in \mathcal{I}_n}[||\hat{\mu}_i(G_{\bm{\theta_g}}(\bm{z})) - \mu_i||_2\\
        + ||\hat{\sigma}_i(G_{\bm{\theta_g}}(\bm{z})) - \sigma_i||_2],
        \label{eq:loss_bns}
    \end{aligned}
\end{equation}
where $\mathcal{I}_n$ is the index set of BN layers in the $n$-th teacher.
Because multiple teachers have learned diverse feature representations from the original dataset \cite{DBLP:journals/corr/abs-2012-09816}, the modified BN statistics constraint takes the representations learned by all teachers into consideration.
Hence, the BN loss can assist the generator in producing more informative samples than simply using a single pretrained teacher, and these samples can help achieve better knowledge transfer.

\textbf{The two-level KD loss} contains two loss terms designed for knowledge distillation.
As mentioned above, we set the number of output headers in the student to be equal to that of pretrained teachers to assign each header a unique teacher.
Thus, we can extend the adversarial distillation framework to the multi-teacher scenario:
\begin{equation}
    \begin{aligned}
        \mathcal{L}_{head}&(\bm{\theta_s},\bm{\theta_g})=\\
        \frac{1}{N}&\sum_{n=1}^{N} \mathbf{E}_{\bm{z} \sim \mathcal{N}(0,\bm{I})}||S_{\bm{\theta_s},n}(G_{\bm{\theta_g}}(\bm{z}))-T_n(G_{\bm{\theta_g}}(\bm{z}))||_1,
        \label{eq:loss_head}
    \end{aligned}
\end{equation}
where $S_{\bm{\theta_s},n}(\cdot)$ denotes the output coming from the $n$-th header of the student.
Note that we replace the KL divergence measurement in the original adversarial distillation framework with the $l_1$ norm measurement, which has been proven to be more effective in data-free KD \cite{DBLP:conf/cvpr/TruongMWP21}.

Taking the average predictions from multiple trained models as the final result is a simple but effective ensemble method for DNN \cite{DBLP:journals/corr/abs-2012-09816}.
We want to preserve the ensemble gain of pretrained teachers in our tiny student model.
Thus, we design an ensemble prediction loss term to guide the ensemble of student headers to output results that resemble those of the teacher ensemble:
\begin{equation}
    \mathcal{L}_{ens}(\bm{\theta_s})= \mathbf{E}_{\bm{z} \sim \mathcal{N}(0,\bm{I})}||\frac{1}{N}\sum_{n=1}^{N} (S_{\bm{\theta_s}}(G(\bm{z}))-T_n(G(\bm{z})))||_1.
    \label{eq:loss_ens}
\end{equation}
The header loss constitutes the adversarial target, and the ensemble loss term is only used to train the student.

\textbf{The feature loss} introduces supervision from intermediate layers of teachers.
As shown in previous works \cite{DBLP:journals/corr/RomeroBKCGB14, DBLP:conf/cvpr/YimJBK17, DBLP:journals/corr/HuangW17a}, knowledge from intermediate layers can also guide the learning of the student.
We adopt this idea and design a feature supervision term.
For simplicity and flexibility, this supervision is only imposed on the last feature extraction layer of the teachers and the student, i.e., layers whose outputs are fed into classifiers.
When a model takes $\bm{x}$ as input, we denote the output of these layers as $f_{t,n}(x)$ and $f_{s,n}(x)$, where $t$ and $s$ denote teachers and student headers, respectively, and $n$ indexes them.
Hence, we can write the feature loss term as:
\begin{equation}
    \mathcal{L}_{feat}(\bm{\theta_s})= \frac{1}{N}\sum_{n=1}^{N} \mathbf{E}_{\bm{z} \sim \mathcal{N}(0,\bm{I})}|| f_{s,n}(G(\bm{z}))-f_{t,n}(G(\bm{z}))||_1.
        \label{eq:loss_feat}
\end{equation}

\textbf{The objective functions} of the generator and the student are composed of the above introduced loss terms.
They are formulated as:
\begin{equation}
    \begin{aligned}
        \mathcal{L}_{s}(\bm{\theta_s})&=\mathcal{L}_{head}(\bm{\theta_s},\bm{\theta_g})+\alpha\mathcal{L}_{ens}(\bm{\theta_s})+\beta\mathcal{L}_{feat}(\bm{\theta_s}),\\
        \mathcal{L}_{g}(\bm{\theta_g})&=-\mathcal{L}_{head}(\bm{\theta_s},\bm{\theta_g})+\gamma\mathcal{L}_{bn}(\bm{\theta_g}),
    \end{aligned}
    \label{eq:loss_main}
\end{equation}
where $\alpha$, $\beta$, and $\gamma$ are balancing hyperparameters.
The asymmetry in our adversarial data-free KD approach is apparent from the above objective functions.
Compared with symmetric adversarial distillation, our approach introduces model-specific optimization objectives and can achieve better performance, which will be shown in our experiments.

\begin{algorithm}[t]
    \caption{Asymmetric adversarial data-free KD.}
    \begin{algorithmic}
    \STATE 
    \STATE \textbf{Notations:} learning rates $\eta_s,\eta_g$; epochs $E_{\text{max}}$; iterations in one epoch $I_{\text{max}}$; student update steps in one iteration $S_{\text{max}}$
    \STATE \textbf{Initialize}$(\bm{\theta_s};\bm{\theta_g})$
    \STATE \textbf{for} epoch \textbf{in} $1:E_{\text{max}}$; \textbf{do}
    \STATE \hspace{0.5cm} \textbf{for} iteration \textbf{in} $1:I_{\text{max}}$; \textbf{do}
    \STATE \hspace{1.0cm} \textit{\# Knowledge Transfer}
    \STATE \hspace{1.0cm} \textbf{for} step \textbf{in} $1:S_{\text{max}}$; \textbf{do}
    \STATE \hspace{1.5cm} $\bm{z}\sim\mathcal{N}(0,\bm{I})$; samples$\gets G(\bm{z})$
    \STATE \hspace{1.5cm} compute $\mathcal{L}_s$ with Eqn. \ref{eq:loss_head}, \ref{eq:loss_ens}, \ref{eq:loss_feat}
    \STATE $$\bm{\theta_s} \gets \bm{\theta_s} - \eta_s \frac{\partial\mathcal{L}_s}{\partial \bm{\theta_s}}$$
    \STATE \hspace{1.0cm} \textbf{end for}
    \STATE \hspace{1.0cm} \textit{\# Discrepancy Maximization}
    \STATE \hspace{1.0cm} $\bm{z}\sim\mathcal{N}(0,\bm{I})$; samples$\gets G(\bm{z})$
    \STATE \hspace{1.0cm} compute $\mathcal{L}_g$ with Eqn. \ref{eq:loss_bns}, \ref{eq:loss_head}
    \STATE $$\bm{\theta_g} \gets \bm{\theta_g} - \eta_g \frac{\partial\mathcal{L}_g}{\partial \bm{\theta_g}}$$
    \STATE \hspace{0.5cm} \textbf{end for}
    \STATE \hspace{0.5cm} decay learning rates $\eta_s,\eta_g$
    \STATE \textbf{end for}
    \end{algorithmic}
    \label{algo:kd}
\end{algorithm}

\textbf{The optimization} of our asymmetric adversarial data-free KD resembles that in the original adversarial distillation approach, which has a two-stage form.
We summarize it in Algorithm \ref{algo:kd}.
We adopt the setting that trains the student more frequently than the generator to keep the adversarial process balanced \cite{dfad}.
Moreover, during the whole training process, we let BN layers in the generator and the student keep collecting the mean and variance of all generated samples.
This setting makes the learning of the generator and the student easy and improves the stability of the adversarial training process because they can track the activation statistics of each other.

\begin{algorithm}[t]
    \caption{Attention-based prediction aggregation.}
    \begin{algorithmic}
    \STATE 
    \STATE \textbf{Notations:} learning rates $\eta_q$
    \STATE \textbf{Initialize}$(\bm{q})$
    \STATE \textbf{while} not converged; \textbf{do}
    \STATE \hspace{0.5cm} \textbf{for} $(\bm{x},y)$ \textbf{in} $(\mathcal{\hat{X}},\mathcal{\hat{Y}})$; \textbf{do}
    \STATE \hspace{1.0cm} $\bm{z}\sim\mathcal{N}(0,\bm{I}),\theta \sim \text{Beta}(1,1)$
    \STATE \hspace{1.0cm} $\hat{\bm{x}}\gets\theta \bm{x} + (1-\theta)G(\bm{z})$
    \STATE \hspace{1.0cm} compute label $y'$ of $G(\bm{z})$ with Eqn. \ref{eq:label}
    \STATE \hspace{1.0cm} compute $\mathcal{L}_{attn}$ with Eqn. \ref{eq:loss_attn}
    \STATE $$\bm{q} \gets \bm{q} - \eta_q \frac{\partial\mathcal{L}_{attn}}{\partial \bm{q}}$$
    \STATE \hspace{0.5cm} \textbf{end for}
    \STATE \textbf{end while}
    \end{algorithmic}
    \label{algo:attn}
\end{algorithm}

\subsection{Attention-based Prediction Aggregation}
\label{subsec:attn}
When some real samples are available, we use an attention module to aggregate predictions of student headers.
The attention module takes a simple key-value attention form, and a Mixup \cite{mixup} of real and generated samples is adopted for training the model.

The architecture of the attention-based prediction aggregation module is illustrated in the upper right corner of Fig. \ref{fig:overview}.
The key vectors are composed of the features before classifiers, and the value vectors consist of logits.
By computing the inner product between the key vectors and a learnable query vector and processing the result with a softmax function, we can obtain a weight vector for computing the weighted sum of all logit vectors.
Then, the weighted logit vector is regarded as the final prediction.
To formulate the above process, we represent key and value vectors by matrices:
$$
\begin{aligned}
    \bm{K}&=[f^T_{s,1}(\bm{x}),f^T_{s,2}(\bm{x}),\dots,f^T_{s,N}(\bm{x})] \in \mathbb{R}^{N \times D},\\
    \bm{V}&=[S_{\bm{\theta_s},1}^T(\bm{x}),S_{\bm{\theta_s},2}^T(\bm{x}),\dots,S_{\bm{\theta_s},N}^T(\bm{x})] \in \mathbb{R}^{N \times C},
\end{aligned}
$$
where $D$ is the dimension of the feature vectors, $C$ is the dimension of the logit vectors, i.e., the number of classes, and $\bm{x}$ is an arbitrary input.
Denoting the query vector as $\bm{q} \in \mathbb{R}^{D \times 1}$, we can then write the weighted result $\hat{S}_{\bm{\theta_s}}(\bm{x})$ as:
\begin{equation}
    \begin{aligned}
        \hat{S}_{\bm{\theta_s}}(\bm{x})&=\bm{V}^T \bm{w}\\
        &=\bm{V}^T \text{softmax}(\frac{\bm{K}\bm{q}}{\sqrt{D}}),
    \end{aligned}
    \label{eq:attn}
\end{equation}
where $\bm{w}$ is the weight vector.

Furthermore, considering that there may exist a domain gap between real and generated samples, which harms the aggregated result, we use a mix of real and generated samples to train the query vector, resembling the Mixup approach \cite{mixup}.
The loss function is defined as:
\begin{align}
    \begin{split}
        \mathcal{L}_{attn}(\bm{q})&=\mathbf{E}_{(\bm{x},y),\bm{z},\theta}[ \theta \mathcal{D}_{CE}(\hat{S}_{\bm{\theta_s}}(\hat{\bm{x}}),y)+\\
        &~~~~~~~~~~~~~~~~(1-\theta)\mathcal{D}_{CE}(\hat{S}_{\bm{\theta_s}}(\hat{\bm{x}}),y')],
        \label{eq:loss_attn}
    \end{split}\\
    \begin{split}
        \hat{\bm{x}}&=\theta \bm{x} + (1-\theta)G(\bm{z}),
        \label{eq:mixed_x}
    \end{split}\\
    \begin{split}
        y'&=\arg \max\frac{1}{N}\sum_{n=1}^{N}T_{n}(G(\bm{z})),
        \label{eq:label}
    \end{split}
\end{align}
$$
\begin{aligned}
    (\bm{x},y)&\in(\mathcal{\hat{X}},\mathcal{\hat{Y}}),\bm{z}\sim\mathcal{N}(0,\bm{I}),\theta \sim \text{Beta}(1,1),
\end{aligned}
$$
where $\hat{\bm{x}}$ is a mixed sample, $y'$ is the label of $G(\bm{z})$ given by the average of teachers, $(\mathcal{\hat{X}},\mathcal{\hat{Y}})$ is an available subset of the original dataset, and $\theta$ is a scalar sampled from a beta distribution, which controls the mixing ratio of real and generated samples.
We summarize the training procedure in Algorithm \ref{algo:attn}.

\section{Experiments}
\label{sec:exp}

\subsection{Datasets and Baselines}
\label{subsec:exp_data}
We evaluate the proposed CDFKD-MFS framework on three public computer vision datasets: CIFAR-100 \cite{cifar}, Caltech-101 \cite{caltech101}, mini-ImageNet \cite{miniimagenet}.
The details of each dataset are listed as follows:
\begin{itemize}
    \item \textbf{CIFAR-100.} CIFAR-100 consists of 60000 images that belong to 100 classes uniformly. Each image is of size $32\times32$. In each class, there are 500 images for training and 100 images for testing.
    \item \textbf{Caltech-101.} Caltech-101 consists of images belonging to 101 categories, whose sizes are approximately $300\times200$. The number of images in each class ranges from 40 to 800. We randomly sample 80\% of the dataset as the training set, and the remaining samples are used as the testing set. Images in this dataset are resized to $128\times128$. 
    \item \textbf{mini-ImageNet.} mini-ImageNet is designed for few-shot learning, which a subset of the ImageNet1k dataset \cite{imagenet}. This dataset consists of 60000 images from 100 classes. We resplit this dataset for classification, where 80\% of images sampled in each class are used for training, and the remaining 20\% of images are used for testing. All images are resized to $224\times224$ in our experiment.
\end{itemize}

We compare the CDFKD-MFS framework with both data-required and data-free counterparts.
We only select KD \cite{kd} as the data-required counterpart for comparison.
Data-free approaches are either generator-based or optimization-based.
In addation to the method in our conference work, we select 3 generator-based methods and 2 optimization methods as the baselines, which are listed as follows:
\begin{itemize}
    \item \textbf{CDFKD \cite{cdfkd}.} The approach in our previous conference work for multi-teacher data-free KD, which is generator-based. We slightly modify this approach by replacing the outdated data generation module with an adversarial-trained generator.
    \item \textbf{DFED \cite{dfed}.} A generator-based approach for multi-teacher data-free KD that uses adversarial distillation and a BN statistics constraint.
    \item \textbf{DFAD \cite{dfad}.} A generator-based approach that uses only adversarial distillation.
    \item \textbf{DFQ \cite{dfq}.} A generator-based approach that adopts adversarial distillation, a BN statistics constraint, and two extra entropy constraints on the sample category.
    \item \textbf{ADI \cite{adi}.} An optimization-based approach that obtains substitute samples in advance and then conducts knowledge transfer with these samples.
    \item \textbf{CMI \cite{cmi}.} An optimization-based approach that adopts contrastive learning \cite{DBLP:conf/icml/ChenK0H20} for diverse sample generation.
\end{itemize}

\subsection{Implementation Details}
\label{subsec:exp_param}
\textbf{Models.} 
There are three models in the proposed framework: pretrained teachers, a student, and a generator.
Similar to existing data-free approaches, we adopt ResNet-34 \cite{resnet} as the teacher network architecture.
The number of teachers is set to $3$.
The only difference between these teachers is their random initialized parameters at the beginning of training, which is proven to be sufficient to obtain diverse teachers \cite{DBLP:journals/corr/abs-2012-09816}.
The backbone of the student is a ResNet-18 network.
We set one feature-sharing point after each residual block group in the backbone, where the total number is $4$.
The number of headers in the student is equal to that of teachers, i.e., $3$.
For all header blocks, the number of their output channels equals the channel number of the shared features with which they are concatenated.
The architecture of the generator comes from DCGAN \cite{dcgan}.
For CIFAR-100, the generator adopts nearest-neighbor upsampling layers with a scale factor equal to $2$ to achieve upsampling.
The output block is a Tanh activation function followed by a BN layer.
For Caltech-101 and mini-ImageNet, the generator adopts transposed convolutional layers with a kernel size equal to $4$ for upsampling.
Output block of the Caltech-101 generator is only a Tanh activation function, while the block of the mini-ImageNet generator is the same as that of the CIFAR-100 generator.
For all datasets, we set the dimension of the normal noise as $256$.

\begin{table}
    \caption{Part of The Training Settings for Asymmetric Adversarial Data-free KD.}
    \begin{center}
        \begin{tabular}{cccc}
        \toprule
        &\textbf{CIFAR-100}&\textbf{Caltech-101}&\textbf{mini-ImageNet}\\
        \hline
        $E_{\text{max}}$&$300$&$300$&$100$\\
        $I_{\text{max}}$&$50$&$50$&$750$\\
        $S_{\text{max}}$&$5$&$5$&$5$\\
        Batch size&$256$&$64$&$64$\\
        $\alpha$&$5$&$5$&$1$\\
        $\beta$&$0.2$&$0.2$&$0.05$\\
        $\gamma$&$0.1$&$0.1$&$0.1$\\
        lr milestone&$[100,200]$&$[100,200]$&$[30,60,90]$\\
        \bottomrule
        \end{tabular}
        \label{tab:param}
    \end{center}
\end{table}

\textbf{Training.} 
Settings for conducting asymmetric adversarial data-free KD on the different datasets are not the same.
We list the different hyperparameters in Table \ref{tab:param}, and introduce the others as follows.
In optimization, the student is trained through stochastic gradient descent with a learning rate $\eta_s$ of $0.1$, momentum of $0.9$, and weight decay of $0.0005$.
The generator is trained with the Adam optimizer \cite{adam}, where the initial learning rate $\eta_g=0.001$, and $\eta_s$ and $\eta_g$ are multiplied by $0.1$ when the running epoch is a member of the lr milestone array.
Furthermore, we use a batch size of $128$ and the AdamW optimizer \cite{adamw} with a learning rate $\eta_q=0.01$ and weight decay $0.0001$ to train the query vector in the attention-based aggregation module.

All baselines are run with their official released code, except DFQ \cite{dfq}, which we adopt the implementation in CMI \cite{cmi}.
For optimization-based approaches, i.e., ADI \cite{adi} and CMI \cite{cmi}, we generate images whose number is equal to the original dataset in advance and then implement the knowledge transfer.
We run each setting $3$ times.

\begin{table*}
    \caption{Knowledge Distillation Results.}
    \begin{center}
        \begin{tabular}{cccccccccc}
        \toprule
        \multirow{2}{*}{Model/Method}&\multirow{2}{*}{Year}&\multirow{2}{*}{Data-required}&\multirow{2}{*}{Params(M)}&\multicolumn{2}{c}{CIFAR-100}&\multicolumn{2}{c}{Caltech-101}&\multicolumn{2}{c}{mini-ImageNet}\\
        \cmidrule(r){5-6} \cmidrule(r){7-8} \cmidrule(r){9-10}
        &&&&FLOPs(G)&Acc(\%)&FLOPs(G)&Acc(\%)&FLOPs(G)&Acc(\%)\\
        \hline
        Teacher&-&$\checkmark$&$21.3$&$1.16$&$78.07$&$1.20$&$78.54$&$3.67$&$78.13$\\
        Teacher ensemble&-&$\checkmark$&$64.0$&$3.48$&$80.40$&$3.60$&$80.83$&$11.01$&$80.99$\\
        Student&-&$\checkmark$&$11.2$&$0.56$&$77.42$&$0.59$&$77.40$&$1.82$&$77.62$\\
        KD \cite{kd}&2015&$\checkmark$&$11.2$&$0.56$&$77.62$&$0.59$&$77.76$&$1.82$&$77.76$\\
        \hline
        DFAD \cite{dfad}&2019&$\times$&$11.2$&$0.56$&$68.35$&$0.59$&$70.60$&$1.82$&$56.51$\\
        DFQ \cite{dfq}&2020&$\times$&$11.2$&$0.56$&$76.14$&$0.59$&$77.80$&$1.82$&$72.94$\\
        ADI \cite{adi}&2020&$\times$&$11.2$&$0.56$&$51.02$&$0.59$&$23.54$&$1.82$&$32.84$\\
        CMI \cite{cmi}&2021&$\times$&$11.2$&$0.56$&$73.05$&$0.59$&$71.39$&$1.82$&$72.87$\\
        DFED \cite{dfed}&2021&$\times$&$11.2$&$0.56$&$75.87$&$0.59$&$78.64$&$1.82$&$74.05$\\
        CDFKD \cite{cdfkd}&2021&$\times$&$20.8$&$0.71$&$66.47$&$0.75$&$78.04$&$2.28$&$59.53$\\
        \hline
        CDFKD-MFS(w/o attention)&-&$\times$&$18.2$&$0.64$&$77.69$&$0.68$&$79.59$&$2.09$&$76.94$\\
        CDFKD-MFS(w attention)&-&$\checkmark(10\%)$&$18.2$&$0.64$&$\mathbf{77.81}$&$0.68$&$\mathbf{79.82}$&$2.09$&$\mathbf{77.04}$\\     
        \bottomrule
        \end{tabular}
        \label{tab:performance}
    \end{center}
\end{table*}

\begin{figure*}
    \centering
    \includegraphics[width=0.9\textwidth]{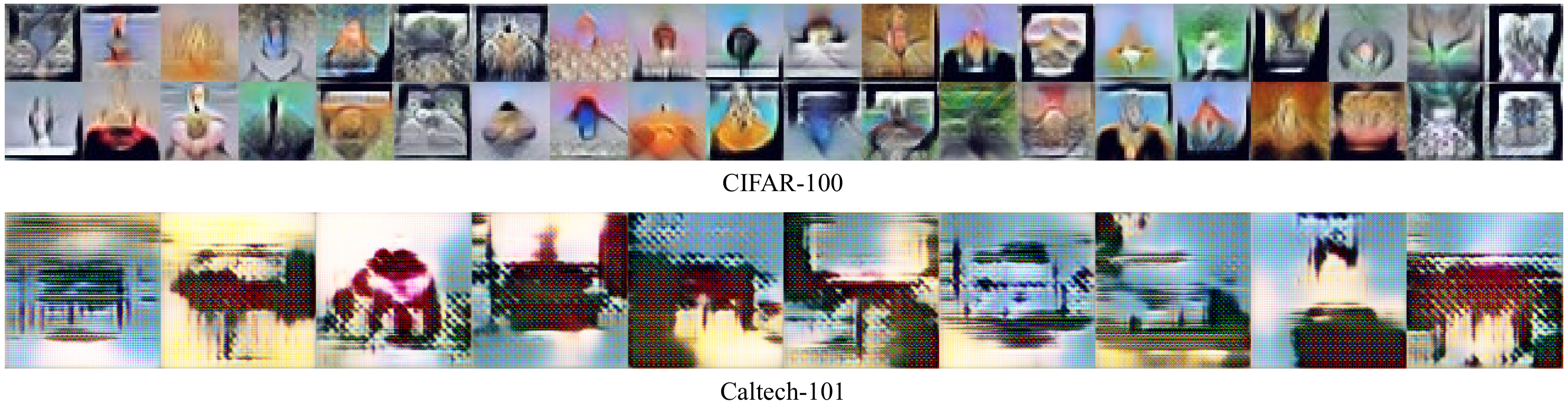}
    \caption{Visualization of images generated on CIFAR-100 (top) and Caltech-101 (bottom). These abstract samples are suitable for knowledge transfer because each one of them contains information from more than one category.}
    \label{fig:vis_cc}
\end{figure*}

\subsection{Results on CIFAR-100 and Caltech-101}
\label{subsec:exp_cc}
We first conduct experiments on CIFAR-100 and Caltech-101, which are usually used by previous data-free counterparts. 
The results are summarized in Table \ref{tab:performance}.

\textbf{Result.} 
We first analyze the results of CIFAR-100.
Given some pretrained teacher models, the ensemble is an effective method to improve performance.
By simply taking their average to obtain the new prediction, we can achieve a more than $2\%$ accuracy improvement.
Although the ensemble method performs well, it requires tremendous computing resources, which is unaffordable for edge devices.
Direct training of a student model or using KD to transfer the knowledge of pretrained teachers into a student model can result in a tiny model applying to deployment.
However, when the original dataset is unavailable, data-free KD approaches are required.
The results show that all data-free methods can transfer the knowledge successfully, and the proposed method performs best among them.
Furthermore, when some original data are available, the performance can be further improven with the attention-based aggregation module.
Our method can achieve at least $1.67\%$ performance improvement compared with the data-free baselines.
Our approach also outperforms the vanilla KD method by $0.19\%$ because multiple pretrained teachers can provide more knowledge than a single teacher.
On Caltech-101, our framework can achieve a $1.18\%$ performance improvement over other data-free counterparts.

\begin{figure*}
    \centering
    \includegraphics[width=0.85\textwidth]{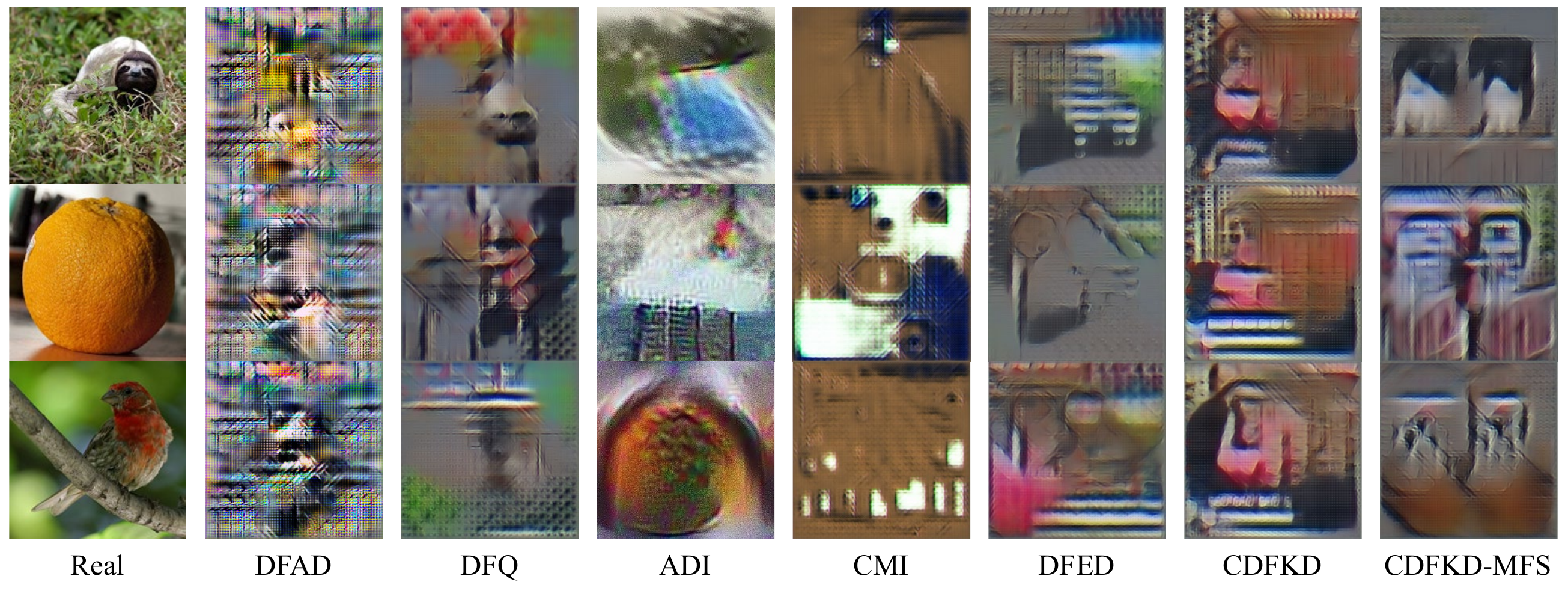}
    \caption{Comparison of images generated on mini-ImageNet. Noting that samples are randomly selected, there is no guarantee that each row belongs to the same category. Horizontally symmetric samples generated by our method are more informative and can assist the student in learning the horizontal flip invariance of teachers.}
    \label{fig:vis_miniimagenet}
\end{figure*}

\textbf{Discussion.} 
According to the results of baselines with unmodified student architecture, a linear single-header student seems insufficient to preserve the diverse knowledge of multiple teachers because there is a noticeable performance gap between itself and the teachers.
Hence, decoupling the knowledge by a multi-header student to bridge the gap is reasonable.
Our previous CDFKD framework adopts this idea but seems ineffective. 
Considering that this unsatisfactory result may come from the inappropriately designed header architecture, we redesign a multi-header student with a multi-level feature-sharing structure, and it performs well. 
We find that the proposed architecture brings more parameters and FLOPs requirement, but different from the apparent increase in the number of parameters, owing to the depthwise separable convolution, the FLOPs requirement increases slightly, whose rate is from $10\%$ to $20\%$.
Compared with cheap storage, computing resources usually constrain the deployment of DNNs more.
With little extra computing resources cost, our method can train a student with performance comparable to a single teacher in a data-free manner.

We also note that the proposed method even outperforms a single pretrained teacher on Caltech-101.
We conjecture that this is because Caltech-101 is a simple dataset with fewer samples, so each teacher fits the training set by directly remembering most of the samples and learns fewer universal feature representations \cite{DBLP:journals/corr/abs-2012-09816}.
The insufficiently learned representations constrain their performance on the testing set.
When multiple pretrained teachers are used, their diverse representations are aggregated into a single student model.
Hence, the student learns more feature representations and performs better than a single teacher.

\begin{figure}
    \centering
    \includegraphics[width=\linewidth]{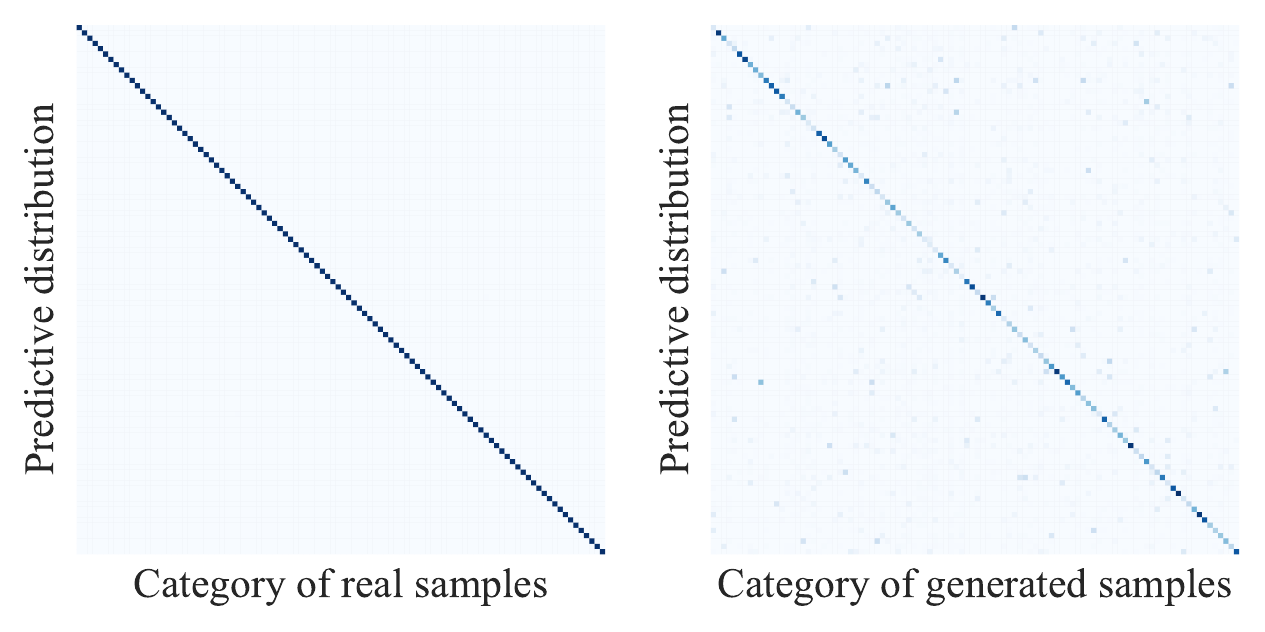}
    \caption{Confusion matrices of real samples (left) and generated samples (right) on CIFAR-100. Generated samples contain more cross-category information than real samples.}
    \label{fig:cm}
\end{figure}

\textbf{Sample visualization.} 
We provide some generated samples of the proposed method in Fig. \ref{fig:vis_cc}.
These samples seem to be abstract.
We conjecture that the cross-category information contained in each sample causes their confusing appearance, which means that each sample belongs to more than one category simultaneously.
To verify this, we show confusion matrices of real samples and generated samples on CIFAR-100 in Fig. \ref{fig:cm}.
The predictive distribution of all samples is obtained with one pretrained teacher model, and the label of each generated sample is set to the index of the max value in the predictive distribution vector.
The confusion matrices illustrate that each generated sample indeed belongs to multiple categories in the view of the teacher.
According to the original KD paper \cite{kd}, these samples containing cross-category information are beneficial for transferring knowledge, with which teachers can teach the student the relations among multiple categories, not only the knowledge within each individual category.

\subsection{Results on mini-ImageNet}
\label{subsec:exp_mi}
We then test the proposed framework on the mini-ImageNet dataset.
Except for ADI, all the data-free baselines were evaluated with at most $128\times128$ input size in their original paper.
We use the mini-ImageNet dataset with a $224\times224$ input size setting to test their performance on larger samples.

\textbf{Result analysis.} 
The right two columns of Table \ref{tab:performance} show the results on mini-ImageNet.
As we can see, the performance gap between students and teachers increases, because this dataset is complex.
Compared with the data-free counterparts, the proposed method can still learn most of the knowledge and achieve at least a $2.99\%$ performance improvement over other methods.
We compare generated images from each method to demonstrate the superiority of the proposed method.

\textbf{Generated sample comparison.} 
We provide $3$ samples generated by each baseline in Fig. \ref{fig:vis_miniimagenet}.
The far left column shows some samples from the original dataset.
The second and third columns are samples in two generator-based single-teacher approaches: DFAD and DFQ.
Without the BN statistics constraint, samples of DFAD contain more repeating texture, which is meaningless compared with DFQ.
The next two columns are samples obtained with optimization-based approaches, which appear to be more like real images than the other methods, especially the samples from ADI.
Nevertheless, these methods require the precollected mean and variance of the original training samples, which may not be available in some more strict data-free scenarios.
Moreover, optimization-based approaches perform worse than the generator-based counterparts in our experiments.
The last three columns show samples generated by the multi-teacher methods.
Samples in DFED are similar to those in DFQ because DFED regards the average of multiple teachers as one teacher.
However, samples in CDFKD have a similar appearance, which implies that mode collapse \cite{gan} occurred in the generator.
Mode collapse means that a generator can only output a single image with a few variations, which is a common problem in GANs.
We conjecture that the tree-like student confused the knowledge in the backbone during backpropagation, which hinders more useful information flowing into the generator.
Samples generated in the proposed CDFKD-MFS framework appear different from the other methods.
More specifically, they seem horizontally symmetric because random horizontal flipping is adopted as one of the data augmentation schemes when training teachers.
This symmetry means that the generator can produce samples containing more teacher knowledge, and hence, the student achieves better performance. 

\subsection{Ablation Study}
\label{subsec:exp_ablation}
To understand each part of the proposed framework, we design ablation studies, where the framework is evaluated on CIFAR-100 with no available sample.
We study the effectiveness of two auxiliary loss terms and the impact of teacher number.

\begin{table}
    \caption{Ablation Study on Auxiliary Loss Terms.}
    \begin{center}
        \begin{tabular}{ccc}
        \toprule
        Ensemble loss&Feature loss&Accuracy(\%)\\
        \hline
        $\times$&$\times$&$73.62$\\
        $\checkmark$&$\times$&$76.84$\\
        $\times$&$\checkmark$&$77.24$\\
        $\checkmark$&$\checkmark$&$77.69$\\
        \bottomrule
        \end{tabular}
        \label{tab:ablation}
    \end{center}
\end{table}

\textbf{Auxiliary loss}.
There are three auxiliary loss terms for training the student and the teacher.
They are asymmetric parts in the objective functions (\ref{eq:loss_main}), controlled by hyperparameters $\alpha$, $\beta$, and $\gamma$.
Because the BN statistics constraint has been proven effective, we only consider the ensemble loss (\ref{eq:loss_ens}) and the feature loss (\ref{eq:loss_feat}) here. 
Table \ref{tab:ablation} shows the evaluation result.
Each auxiliary loss can help improve the performance, and the best result is achieved when both  terms are adopted.
The improvement is $4.07\%$ when both losses are used.

\begin{figure}
    \centering
    \includegraphics[width=0.8\linewidth]{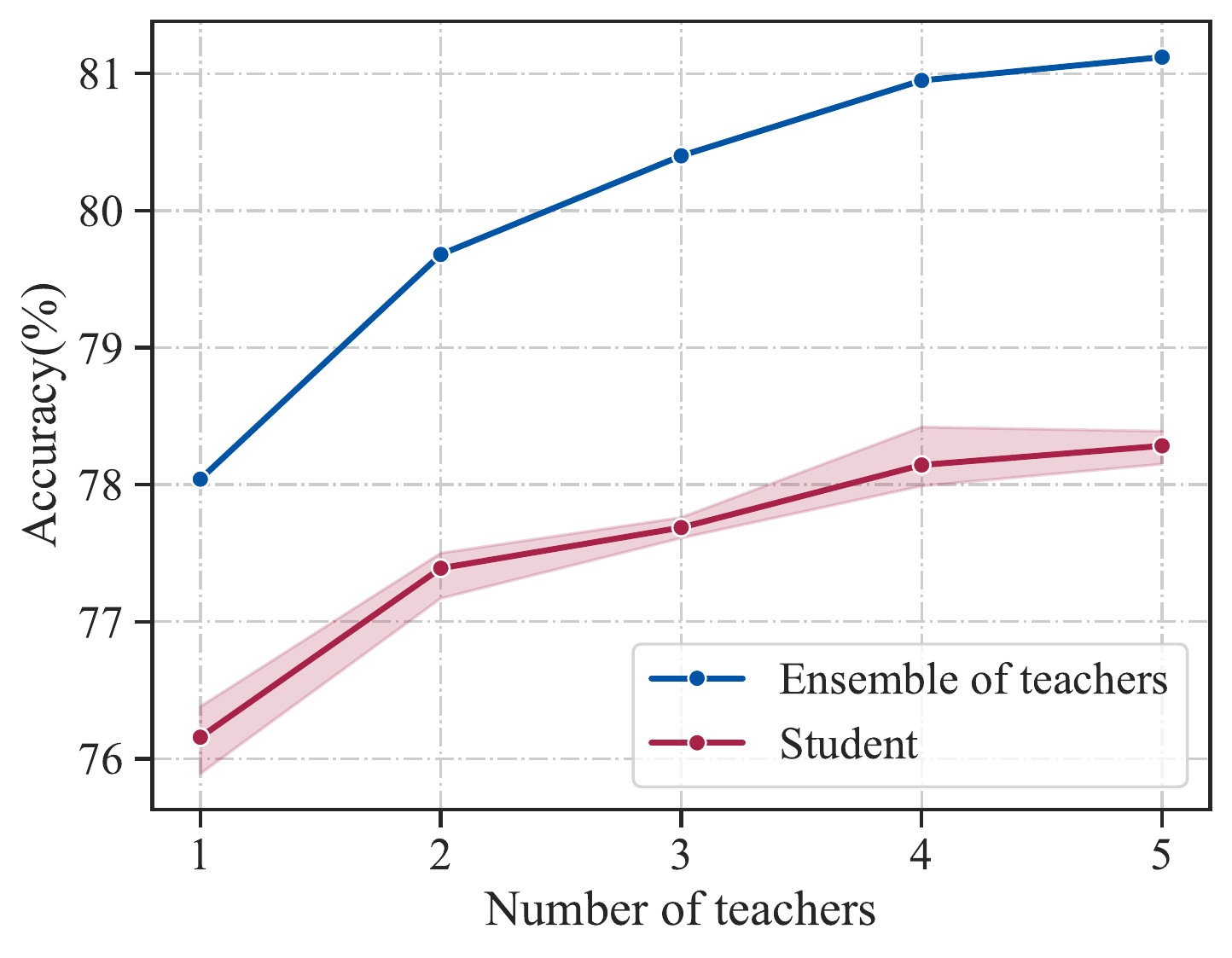}
    \caption{Impact of teacher number. As the teacher number increases, the performance of the student increases as well.}
    \label{fig:ablation}
\end{figure}

\textbf{Impact of teacher number}.
The ensemble method is an effective approach to improve model performance.
We vary the number of teachers in the ensemble from $1$ to $5$ and evaluate how the student performs. 
The result is shown in Fig. \ref{fig:ablation}.
Not surprisingly, the accuracy of the student increased with the number of teachers.
Moreover, as the teacher number increases, the redundant part of the knowledge increases.
Hence, the performance growth rate decreases.

\begin{table*}
    \caption{Knowledge Distillation Results on the CIFAR-100 Dataset with WRN Models.}
    \begin{center}
        \begin{tabular}{cccccccccccc}
        \toprule
        \multicolumn{3}{c}{Teacher}&\multicolumn{3}{c}{Student}&\multicolumn{6}{c}{Accuracy(\%)}\\
        \cmidrule(r){1-3} \cmidrule(r){4-6} \cmidrule(r){7-12}
        Model&Params(M)&FLOPs(M)&Model&Params(M)&FLOPs(M)&T.$^*$&S.$^*$&KD$^*$&DFAD&DFED&CDFKD-MFS\\
        \hline
        ResNet-18&$11.22$&$557$&WRN-40-2&$2.26/2.71$&$329/344$&$77.42$&$75.77$&$75.90$&$45.05$&$69.15$&$\textbf{70.26}$\\
        WRN-40-2&$2.26$&$329$&ResNet-18&$11.22/18.16$&$557/639$&$75.77$&$77.42$&$77.91$&$49.86$&$74.58$&$\textbf{75.80}$\\
        WRN-40-2&$2.26$&$329$&WRN-16-2&$0.70/1.16$&$102/116$&$75.77$&$71.67$&$72.02$&$30.11$&$60.76$&$\textbf{69.08}$\\
        WRN-40-2&$2.26$&$329$&WRN-16-1&$0.18/0.31$&$27/31$&$75.77$&$65.86$&$65.98$&$10.54$&$41.06$&$\textbf{57.57}$\\
        \bottomrule
        \end{tabular}
        \begin{flushleft}
            The left and the right of a slash ($/$) are data of a linear student and a multi-header student, respectively. Data-required methods are marked by an asterisk ($*$).
        \end{flushleft}
        \label{tab:more}
    \end{center}
\end{table*}

\subsection{Results on More Lightweight Models}
\label{subsec:exp_light}
DFED has shown that it is challenging to perform data-free KD on small models, e.g., wide residual networks (WRNs) \cite{wrn}, because it results in remarkable accuracy loss.
To further verify the effectiveness, we compare the proposed framework with previous works on CIFAR-100 with WRN models.
All hyperparameters are the same as those in Section \ref{subsec:exp_cc}, and the attention module is not used.

The results are reported in Table \ref{tab:more}.
We first take a ResNet-18 network as the teacher and a WRN-40-2 network as the student, which contains much fewer parameters than the teacher.
Compared with data-required methods, there is an apparent performance degradation when applying data-free methods, larger than $5\%$.
We then exchange the teacher and the student and observe a smaller performance gap of only approximately $2\%$.
Hence, we conjecture that performing data-free KD gets more challenging as the student becomes smaller.
We take two WRN models with fewer parameters as the student and a WRN-40-2 model as the teacher.
The results follow our conjecture, where the performance gap between data-free and data-required methods becomes more significant when the student becomes smaller.
However, our framework can significantly narrow the performance gap.
When we take the WRN-16-2 model as the student, the proposed method achieves an $8.32\%$ performance improvement compared with the data-free counterparts, and it achieves a $16.51\%$ performance improvement when we use a smaller WRN-16-1 network as the student model.

\subsection{Attention Map Comparison}
\label{subsec:exp_attn}
We further compare attention maps generated by different teachers and student headers on the same sample to study what the student has learned.
Attention maps are obtained on the last residual block of each model with GradCAM \cite{gradcam}.
Samples for comparison come from the mini-ImageNet dataset.

\begin{figure*}[!t]
    \centering
    \includegraphics[width=0.95\textwidth]{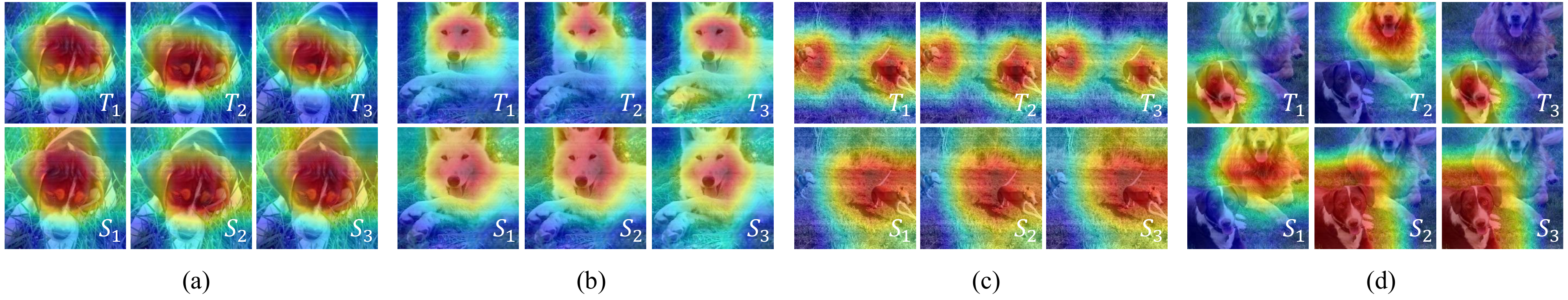}
    \caption{Teacher-student attention map comparison: (a) single object with similar teacher attention maps; (b) single object with different teacher attention maps; (c) multiple objects with similar teacher attention maps; and (d) multiple objects with different teacher attention maps. The lower right corner in each image shows which teacher or student header generates the attention map.}
    \label{fig:supplementary}
  \end{figure*}

Attention maps are shown in Fig. \ref{fig:supplementary}.
We select four different samples according to the number of objects in the sample and the similarity of teacher attention maps.
As we can see, when there is only one object in the sample, the student learns from teachers well.
Each student header gives attention maps similar to its corresponding teacher (Fig. \ref{fig:supplementary}a, \ref{fig:supplementary}b).
However, when multiple objects exist in a single sample, the result is different.
The student either merges distinct teacher attention regions (Fig. \ref{fig:supplementary}c), or confuses attention from different teachers (Fig. \ref{fig:supplementary}d).
Hence, we conjecture that samples containing multiple objects belonging to the label category are the main obstacle for data-free KD.
In our future work, we will study how to guide a student to learn from teachers on such samples.

\section{Conclusion}
\label{sec:conclusion}
In this paper, we propose a framework termed CDFKD-MFS for multi-teacher data-free KD.
The framework consists of a multi-level feature-sharing multi-header student model architecture, an asymmetric adversarial data-free KD algorithm, and an attention-based prediction aggregation module.
We conduct extensive experiments to evaluate our framework.
The results show that in addition to the widely used CIFAR-100 dataset and Caltech-101 dataset, the proposed framework also outperforms other data-free counterparts on the mini-ImageNet dataset under the more challenging $224\times224$ input size.
When some real samples are available, our method achieves better performance by using the attention mechanism to adaptively aggregate predictions of the multi-header student.
Moreover, when the student model is extremely small, the proposed framework can still achieve satisfactory performance, which is difficult for existing works.
In the future, we will explore how to teach a student in data-free KD to perform like the teachers on samples containing multiple objects and further improve performance.

\bibliographystyle{IEEEtran}
\bibliography{ref}

\end{document}